\pdfoutput=1

\documentclass[11pt]{article}

\usepackage{ACL2023}

\usepackage{times}
\usepackage{latexsym}
\usepackage[T1]{fontenc}
\usepackage{multirow}
\usepackage[utf8]{inputenc}

\usepackage{microtype}

\usepackage{inconsolata}
\usepackage{graphicx}
\usepackage{hyperref}       
\usepackage{url}            
\usepackage{booktabs}       
\usepackage{amsfonts}       
\usepackage{nicefrac}       
\usepackage{microtype}      
\usepackage{xcolor}         
\usepackage{graphicx}
\usepackage{float}
\usepackage{subfigure}
\usepackage{amsmath}
\usepackage{bbm}
\usepackage{mfirstuc}
\usepackage{multirow}
\usepackage{color}
\usepackage{wrapfig}
\usepackage{natbib}
\usepackage{textcomp, gensymb}
\usepackage{tablefootnote}
\usepackage[mathscr]{euscript}
\usepackage{diagbox}
\newcommand{\name}{Batch-ICL}
\title{\name: Effective, Efficient, and Order-Agnostic In-Context Learning}

\author{Kaiyi Zhang$^{1}$\thanks{\ \ Equal contribution.},\quad Ang Lv$^{1}$\footnotemark[1],\quad Yuhan Chen$^{1}$,\\ \textbf{Hansen Ha}$^{2}$, \quad \textbf{Tao Xu}$^{2}$, \quad \textbf{Rui Yan}$^{1,3}$\thanks{\ \ Corresponding author: Rui Yan (ruiyan@ruc.edu.cn)} \\
  $^1$Gaoling School of Artificial Intelligence, Renmin University of China\\
  $^2$Ant Group\\
  $^3$Engineering Research Center of Next-Generation Intelligent Search and Recommendation, \\Ministry of Education\\
  \texttt{\{kyzhang02\}@gmail.com}, \texttt{\{anglv,yuhanchen,ruiyan\}@ruc.edu.cn} \\
  \texttt{\{hahansen.hhs,tomas.xt\}@antgroup.com}
}

\begin{document}
\maketitle
\begin{abstract}
In this paper, by treating in-context learning (ICL) as a meta-optimization process, we explain why LLMs are sensitive to the order of ICL examples. This understanding leads us to the development of Batch-ICL, an effective, efficient, and order-agnostic inference algorithm for ICL. Differing from the standard N-shot learning approach, Batch-ICL employs $N$ separate 1-shot forward computations and aggregates the resulting meta-gradients. These aggregated meta-gradients are then applied to the forward computation of a zero-shot query to generate the final prediction. This batch processing approach renders the LLM agnostic to the order of ICL examples. Through extensive experiments and analysis, we demonstrate that Batch-ICL consistently outperforms most permutations of ICL examples. In some cases, it even exceeds the performance of the best order for standard ICL, all while reducing the computational resources required. Furthermore, we develop a novel variant of Batch-ICL featuring multiple ``epochs'' of meta-optimization. This variant implicitly explores permutations of ICL examples, further enhancing ICL performance.\footnote{
Our code is available at \url{https://github.com/Cardinalere/Batch-ICL}.}
\end{abstract}

\section{Introduction}
\citet{brown2020language} demonstrate the capacity of large language models (LLMs) to perform in-context learning (ICL) wherein the input context comprises a handful of illustrative instances of specific tasks. 
In this few-shot setting, LLMs are capable of identifying the task and adapting their response format and domain accordingly.
For instance, when presented with context such as ``I love this movie. Sentiment: positive. \textbackslash n I hate this movie. Sentiment: negative. \textbackslash n This film is interesting. Sentiment:,'' the LLM might accurately recognize the sentiment classification task and provide the appropriate response, which in this case would be ``positive.''
Without training or fine-tuning, ICL sometimes even matches the performance of supervised trained models.

Numerous studies~\citep{olsson2022context, wang-etal-2023-label, dai-etal-2023-gpt, pmlr-v202-li23l, akyurek2023what, pmlr-v202-von-oswald23a, ren2023incontext, xie2022an,lv2024interpreting} have contributed to understanding the mechanism of ICL. 
Specifically, some research~\citep{dai-etal-2023-gpt, ren2023incontext} describes ICL as a meta-optimization where an LLM is utilized as a meta-optimizer.
Meta-gradients are produced through forward processing with ICL examples.
These meta-gradients are then applied to the language model through the attention mechanism, resulting in an effective ICL model.

We propose that these insights shed light on a commonly recognized issue: the ICL capacity of an LLM is highly influenced by the order of examples. 
As emphasized by~\citep{lu-etal-2022-fantastically}, changing the order of ICL examples can lead to significantly different outcomes.
This paper offers a preliminary explanation for this phenomenon:
In an $N$-shot ICL process, the meta-gradient is formed based on $N$ sequentially presented examples.
Due to the causal attention mechanism in LLMs, the meta-gradient shaped by any given example is indirectly affected by the ones that came before it. 
Consequently, the order of these examples plays a critical role in shaping the final ICL model.
This process is similar to training a neural network with $N$ samples where the batch size is one. 
In such cases, the gradients generated by each sample are influenced not only by the sample itself but also by the parameter updated by preceding samples.
The sample order leads to variations in parameters and the overall performance.

Given that gradients from each sample represents a local optimum and is susceptible to causing suboptimal results, we suggest there is much potential for improving standard ICL.
In this paper, we introduce \name, an effective, efficient and order-agnostic inference algorithm for in-context learning. 
Our method diverges from the standard ICL, which typically employs a single \(N\)-shot process for an \(N\)-shot ICL task. 
Instead, we implement \(N\) separate 1-shot forward computations. 
This is then followed by the aggregation of $N$ corresponding meta-gradients at a specific layer. 
These aggregated meta-gradients are then applied to the LLM in the same layer during the forward computation of a zero-shot query, ultimately generating the final predictions.
This is equivalent to increase the meta-batch size in ICL from 1 to $N$, thereby reducing the randomness of meta-optimization and obtaining a better ICL model.
\name\ yields three key advantages:

(1) \name\ alleviates concerns related to the order of examples in ICL. 
Across a variety of tasks, \name\ consistently exhibits improved accuracy compared to the average accuracy achieved through all permutations of ICL examples.  
It sometimes outperforms the best order.
Meanwhile, \name\ reduces the computational resources needed for executing an ICL sequence.

(2) Additionally, we have discovered that although the standard ICL exhibits instability when presented with different ordered sequences, there are advantages to be gained from the interaction among sequential examples.
We expand \name\ into a ``multi-epoch'' variant, which implicitly enumerates the order permutations in a much more efficient manner, leading to further enhancements.

(3) In \name, there is no limit to the number of demonstration examples provided that the length of each example does not surpass the pretrained context length.
This effectively overcomes a significant constraint encountered in standard $N$-shot ICL.

\section{Understanding In-Context Learning from a Meta-Optimization Perspective}
\label{sec:background}

Many works~\citep{1964, Irie2022TheDF, dai-etal-2023-gpt, ren2023incontext} have demonstrated the similarity between the linear attention and the linear layers optimized by gradient descent.
In this section, we briefly review this similarity and then introduce the in-context learning from a meta-optimization perspective.

\subsection{Dual Form between Linear Attention and Gradient Descent in Linear Layers} 
Consider a linear layer defined as: 
\[f(\boldsymbol{x}) = \boldsymbol{W}_{0} \boldsymbol{x},\] 
where \(\boldsymbol{W}_{0} \in \mathbb{R}^{d_{out} \times d_{in}}\) represents the initial weight matrix. 
Given a sequence of input vectors \(\boldsymbol{x}_i \in \mathbb{R}^{d_{in}}, i \in [1, N]\), and their corresponding labels \(\boldsymbol{y}_i \in \mathbb{R}^{d_{out}}\), the error signal \(\boldsymbol{e}_i\) is produced by backpropagation, where \(\boldsymbol{e}_i = -\eta \frac{\partial \mathcal{L}}{\partial y_i}\), with \(\eta\) as the learning rate and \(\mathcal{L}\) as the loss function. 
The weight matrix updates as follows:
\begin{equation}
    \boldsymbol{W}' = \boldsymbol{W}_{0} + \Delta \boldsymbol{W} = \boldsymbol{W}_{0} + \sum^{N}_{i} \boldsymbol{e}_i \otimes \boldsymbol{x}_i.
\end{equation}

Recap that a linear attention is formulated as:
\begin{equation}
    \texttt{LA}(\boldsymbol{V}, \boldsymbol{K}, \boldsymbol{q}) = \boldsymbol{V}^{\top}\boldsymbol{K}\boldsymbol{q} = \sum_{i} \boldsymbol{v}_i (\boldsymbol{k}^{\top}_i \boldsymbol{q}).
\end{equation}

When focusing on the current input $\boldsymbol{x}_N$, we can derive the dual form between the linear layer optimized by gradient descent and the linear attention:
\begin{equation}
    \begin{aligned}
        f(\boldsymbol{x}_{N}) &= (\boldsymbol{W}_{0} + \Delta \boldsymbol{W}) \boldsymbol{x}_{N} \\
        &= \boldsymbol{W}_{0}\boldsymbol{x}_{N} + \sum^{N-1}_{i} (\boldsymbol{e}_i \otimes \boldsymbol{x}_i) \boldsymbol{x}_{N}\\
        &= \boldsymbol{W}_{0}\boldsymbol{x}_{N} + \sum^{N-1}_{i} \boldsymbol{e}_i (\boldsymbol{x}^{\top}_i \boldsymbol{x}_{N})\\
        &= \boldsymbol{W}_{0}\boldsymbol{x}_{N} + \texttt{LA}(\boldsymbol{E}, \boldsymbol{X}_{1:N-1}, \boldsymbol{x}_{N}).
    \end{aligned}
    \label{eq:dual-form}
\end{equation}

\begin{figure*}[t]
    \centering
    \includegraphics[width=0.85\linewidth]{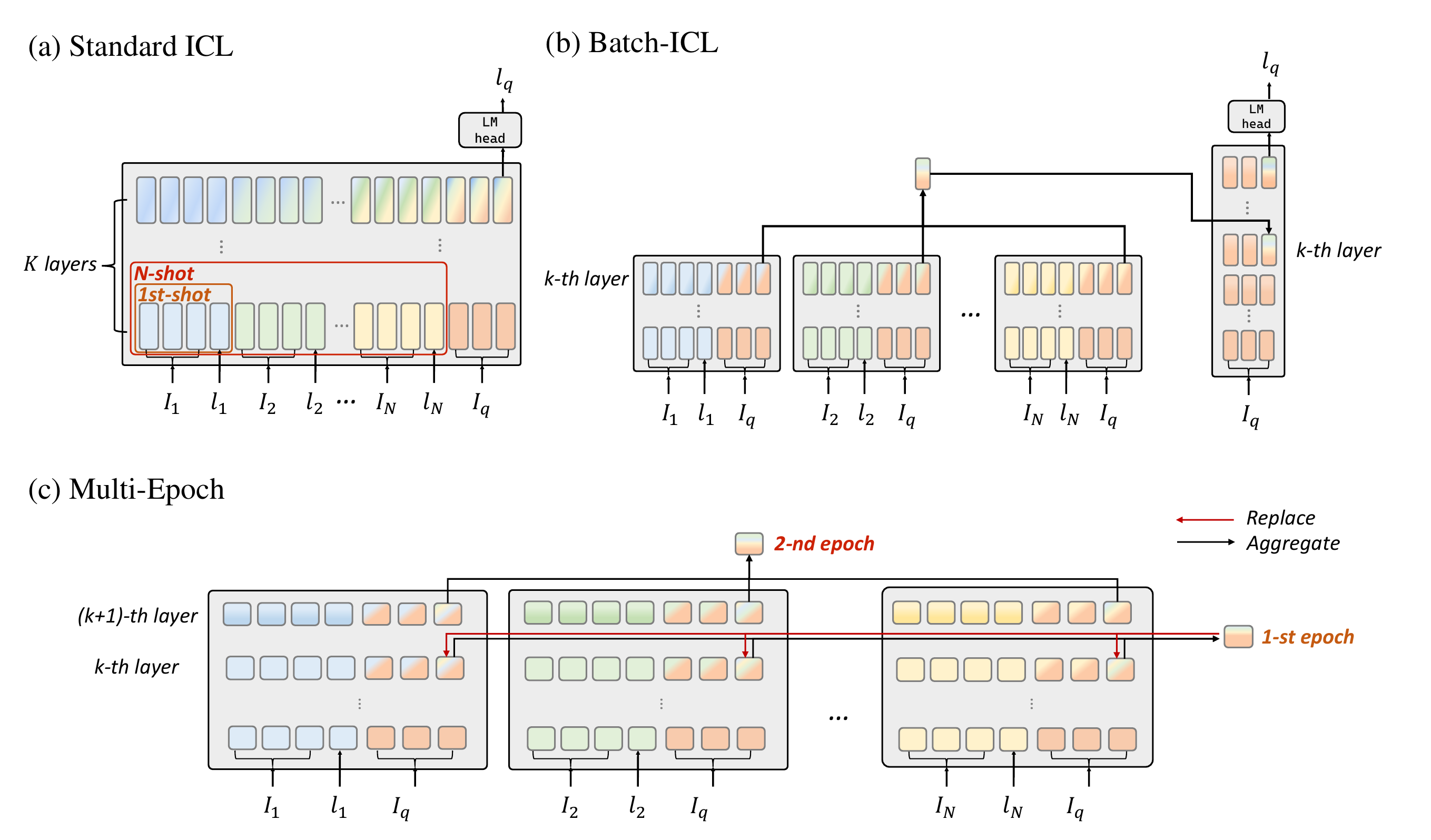}
    \caption{(a) Standard in-context learning. 
    (b) \name\ aggregates the meta-gradients generated during individual 1-shot learning forward computations and applies them to a zero-shot forward process.
    (c) Multi-epoch \name\ further enhances ICL performance, shown here with a 2-epoch overview.
    }
    \label{fig:overview}
\end{figure*}
\section{Method}
\label{sec:method}

\subsection{ICL is an Meta-Optimization}
In an $N$-shot ICL setting, a Transformer~\citep{NIPS2017_3f5ee243}-based LLM consists of $K$ layers.
The LLM processes an input sequence in the form of \(I_1, l_1, I_2, l_2, \dots, I_N, l_N, I_q\), where $I_i$ and $l_i$ together represent a demonstration example, with $I_i$ being the input and $l_i$ as the corresponding label. 
Here, $I_q$ denotes the genuine query, and the LLM's task is to predict its label. 
We represent the embedding of the input sequence at $k$-layer as $\boldsymbol{X}^{k} = [\boldsymbol{x}^{k}_1, \boldsymbol{x}^{k}_2, \dots, \boldsymbol{x}^{k}_q]$ with a special focus on $\boldsymbol{x}^{k}_q$, which represents the last query token's representation. 
Without ambiguity, we will omit layer superscripts.
The output of an attention head can be expressed as follows:
\begin{equation}
\resizebox{0.92\linewidth}{!}{
$f(\boldsymbol{x}_q) = \boldsymbol{W}_V \boldsymbol{X} \ \texttt{Softmax}\left(\frac{(\boldsymbol{W}_K \boldsymbol{X})^T \boldsymbol{W}_Q \boldsymbol{x}_q}{\sqrt{d_{out}}}\right),$}
\end{equation}
where \( \boldsymbol{W}_Q \), \( \boldsymbol{W}_K \), and \( \boldsymbol{W}_V \) are projection matrices belonging to \( \mathbb{R}^{d_{out} \times d_{in}} \). 
These matrices are utilized to compute the attention queries, keys, and values, respectively. 
To simplify the notation below, we denote \( \boldsymbol{W}_Q \boldsymbol{x}_q \) as \( \boldsymbol{q} \) and partition \( \boldsymbol{X} \) into \( [\boldsymbol{X}', \boldsymbol{x}_{q}] \).

~\citet{dai-etal-2023-gpt} describe ICL as a meta-optimization process, inspired by the previously mentioned ``dual form.''
In their framework, the LLM acts as the meta-optimizer, with the meta-gradient according to demonstrations generated during the forward computation and applied to the model through the attention mechanism:
\begin{equation}
\begin{aligned}
f(\boldsymbol{x}_q) & \approx \boldsymbol{W}_V [\boldsymbol{X}'; \boldsymbol{x}_q] (\boldsymbol{W}_K [\boldsymbol{X}'; \boldsymbol{x}_q])^{\top} \boldsymbol{q}\\ 
& = \underbrace{\boldsymbol{W}_V \boldsymbol{x}_q (\boldsymbol{W}_K \boldsymbol{x}_q)^{\top}}_{\text{Denoted as $\boldsymbol{W}_{0}$}} \boldsymbol{q} \ + \\
& \ \ \underbrace{\texttt{LA}(\boldsymbol{W}_V \boldsymbol{X}', \boldsymbol{W}_K \boldsymbol{X}', \boldsymbol{q})}_{\text{$N$ demonstrations' effect, denoted as $\Delta \boldsymbol{W}_{N} \boldsymbol{q}$}} \\
& = (\boldsymbol{W}_{0} + \Delta \boldsymbol{W}_{N}) \ \boldsymbol{q}.
\end{aligned}
\label{eq:icl-meta}
\end{equation}

In the equation above, the standard attention is approximated to a relaxed linear attention.
\( \boldsymbol{W}_V \boldsymbol{x}_q (\boldsymbol{W}_K \boldsymbol{x}_q)^{\top} \) is denoted as \( \boldsymbol{W}_{0} \) because \( \boldsymbol{W}_{0} \boldsymbol{q} \) is the attention output in the 0-shot setting. 
According to Eq.~\ref{eq:dual-form}, the outputs from linear attention can be interpreted as the effect of \( N \) demonstrations.
Consequently, this term is denoted as \( \Delta \boldsymbol{W}_{N} \boldsymbol{q} \). 

In this section, we start by proposing an explanation ($\S$\ref{sec:method-explain}) about the sensitivity of LLMs to the order of ICL demonstration examples, which is built upon our comprehension of ICL as a meta-optimization process.
Following this, in $\S$\ref{sec:method-method}, we introduce our proposed solution, named ``\name,'' in which the LLM performs an $N$-shot ICL by processing ICL examples in a ``batch.'' 
In $\S$\ref{sec:method-epoch}, we elaborate on the extension of \name\ to incorporate multiple ``epochs,'' thus fully harnessing its potential.

\subsection{The Order of ICL Examples Affects Model Outputs}
\label{sec:method-explain}
From this meta-optimization perspective, we provide an explanation for the well-known issue of LLMs' sensitivity to the order of ICL examples~\citep{lu-etal-2022-fantastically}. 

Considering the term \(\Delta \boldsymbol{W}_{N}\) in Eq.~\ref{eq:icl-meta}, the causal attention in LLMs permits only later tokens to attend to earlier ones. 
This implies that altering the order of demonstration examples will correspondingly change the content of \(\boldsymbol{X}'\), thereby affecting the meta-gradients \(\Delta \boldsymbol{W}_{N}\) and ultimately the resulting different obtained ICL model and its outputs.
We would like to compare this meta-optimization process to optimizing a neural network: for \(N\) training samples with a batch size of one, it requires \(N\) back-propagation steps. 
During each step, the gradients are computed and iteratively update the model. 
The gradients for each training sample depend not only on the sample itself but also on the parameters updated by preceding samples. 
Therefore, the order of training examples influences the final model parameters and, consequently, its overall performance.

Despite the significant impact of the orders on performance, this explanation also applies to the observation by~\citet{lu-etal-2022-fantastically} that an increase in the number of examples is correlated with greater variance in performance. This occurs because a larger set of examples raises the upper limit of potential performance. However, due to the meta-gradients being computed sequentially from individual samples, the optimization process is directed towards diverging outcomes, most of which are suboptimal.

\subsection{ICL Examples Batching and Meta-Gradients Aggregation}
\label{sec:method-method}

Inspired by the analogy above, we develop an ICL inference algorithm named ``\name'' which empowers LLMs to handle each ICL example and produce the corresponding meta-gradients individually, mirroring the way they are arranged in a training batch. 
By aggregating these meta-gradients and then applying them to a zero-shot forward process, we have observed that LLMs exhibit superior performance compared to standard ICL.

We adhere to the terminology used in Section~\ref{sec:background}. 
Considering a set of $N$ demonstration examples, we opt for the LLM to undertake $N$ individual 1-shot ICL learning processes, diverging from the standard single $N$-shot learning. 
Each sample received by LLMs is formatted as ``$I_{i}, l_{i}, I_{q}$'', and $i$ spans the entire set $N$. 
At a selected layer $k$, we collect $f_{i}(\boldsymbol{x_q})$ at the last position, i.e., the result of Eq.~\ref{eq:icl-meta}.
Since $\boldsymbol{X}^{'}$ in Eq.~\ref{eq:icl-meta} now represents the representation of ``$I_{i}, l_{i}$'', as opposed to the entire set of ICL examples, we accordingly adjust the formula to: $f_{i}(\boldsymbol{x_q}) = (\boldsymbol{W}_{0,i} + \Delta \boldsymbol{W}_{1,i}) \boldsymbol{q}$.

When aggregating these hidden states $[f_{i}(\boldsymbol{x_q})]^{N}_{i=1}$ using the arithmetic mean, we actually aggregate meta-gradients for each 1-shot learning:
\begin{equation}
\begin{aligned}
\Bar{f}(\boldsymbol{x_q}) &= \frac{1}{N} \sum_{i=1}^{N} f_{i}(\boldsymbol{x_q}) \\
&= \frac{1}{N} \sum_{i=1}^{N} (\boldsymbol{W}_{0,i} + \Delta \boldsymbol{W}_{1,i})\boldsymbol{q} \\
& \approx (\boldsymbol{W}_{0} + \frac{1}{N} \sum_{i=1}^{N} \Delta \boldsymbol{W}_{1,i})\ \boldsymbol{q}.
\end{aligned}
\end{equation}
The approximation here is due to the minor variations between $\boldsymbol{W}_{0,i}$ and the actual zero-shot weight $\boldsymbol{W}_{0}$, influenced by $x_q$ being affected by the preceding 1-shot example.

Next, we substitute the zero-shot outputs $\boldsymbol{W}_0 \boldsymbol{q}$ at the same layer $k$ with the aggregated $\Bar{f}(\boldsymbol{x_q})$, to obtain the final prediction.

To determine the value for parameter $k$, taking into account the variability introduced by model size, parameters, and tasks, we employ a general approach: For any given task and LLM, we evaluate $k$ across the range of maximum layers. 
We then choose the $k$ that yields the best performance on the validation set. 
The selected $k$ remains constant during the testing phase for this LLM and task.

We name this inference algorithm as ``\name'' because it is akin to computing gradients during the optimization of a neural network using a batch of inputs.

\subsection{Expanding to Multiple ``Epochs''}
\label{sec:method-epoch}
We notice that the meta-optimization perspective can be extended to encompass ``multiple epochs,'' which in turn further enhances \name. 
To clarify this concept, let's start by distinguishing the layers with superscripts $k$ in Eq.~\ref{eq:icl-meta}:
\begin{equation}
\begin{aligned}
f^{k+1}(\boldsymbol{x_q}^{k+1}) &= f^{k+1}(f^{k}(\boldsymbol{x_q}^{k})). \\
\end{aligned}
\end{equation}
Here, we simplify the feed-forward layer by treating it as a linear transformation, considering it as a unit matrix for clarity purposes.
When we substitute the aggregated $\Bar{f}^{k}(\boldsymbol{x_q}^{k})$ for its individual components $f^{k}_{i}(\boldsymbol{x_q}^{k})$, which are the separate outputs of each 1-shot ICL at layer $k$, and then derive the attention outputs from each $k+1$ layer for further aggregation, we actually engage in a form of meta-optimization during an additional epoch.
Formally, the outputs of each 1-shot ICL at layer $k+1$ now turns to:
\begin{equation}
\resizebox{0.92\linewidth}{!}{$
\begin{aligned}
&f_{i}^{k+1}(\Bar{f}^{k}(\boldsymbol{x_q}^{k})) \\
&= (\boldsymbol{W}_0^{k+1}+\Delta \boldsymbol{W}_{i}^{k+1}) (\boldsymbol{W}_0^{k} +\frac{1}{N}\sum_{j=1}^{N}\Delta \boldsymbol{W}_{j}^{k})\boldsymbol{q} \\
&=(\boldsymbol{W}_0^{k+1}+\Delta \boldsymbol{W}_{i}^{k+1})\frac{1}{N}\sum_{j=1}^{N}(\boldsymbol{W}_0^{k} +\Delta \boldsymbol{W}_{j}^{k})\boldsymbol{q}\\
&=\frac{1}{N}\sum_{j=1}^{N}(\boldsymbol{W}_0^{k+1}+\Delta \boldsymbol{W}_{i}^{k+1})(\boldsymbol{W}_0^{k} +\Delta \boldsymbol{W}_{j}^{k}) \boldsymbol{q}.
\end{aligned}$}
\end{equation}
Notice the expression \((\boldsymbol{W}_0^{k+1}+\Delta \boldsymbol{W}_{i}^{k+1})(\boldsymbol{W}_0^{k}+\Delta \boldsymbol{W}_{j}^{k})\boldsymbol{q}\), which can be understood as conducting meta-optimization for example \(j\) in the first \(k\) layers, followed by a similar meta-optimization for example \(i\) at layer \(k+1\). 
We term this procedure an extra ``epoch'' within the \name, and it can readily be extended to multi-epoch.
Figure~\ref{fig:overview}(c) illustrates the 2-epoch expansion of \name.

Essentially, the multi-epoch \name\ implicitly enumerates all permutations of ICL examples, yet it achieves this more efficiently. 
To understand this, we formulate \(f_{i}^{k+1}(\boldsymbol{x_q}^{k+1})\) as:
\begin{equation}
\begin{aligned}
f_{i}^{k+1}(\boldsymbol{x_q}^{k+1}) = \frac{1}{N}\sum_{j=1}^{N}(\boldsymbol{W}_0^{k+1}+\Delta \boldsymbol{W}_{ji}^{k})\boldsymbol{q} \\
= (\boldsymbol{W}_0^{k+1}+\frac{1}{N} \sum_{j=1}^{N} \Delta \boldsymbol{W}_{ji}^{k})\boldsymbol{q},
\end{aligned}
\end{equation}
where \(\Delta \boldsymbol{W}_{ji}^{k}\) denotes the meta-gradient produced through optimizing example \(j\) in the first \(k\) layers, followed by optimization of example \(i\) in the \(k+1\) layer.
By aggregating \(f_{i}^{k+1}(\boldsymbol{x_q}^{k+1})\) across \(i=1\cdots N\), we derive:
\begin{equation}
\begin{aligned}
\Bar{f}_{i}^{k+1}(\boldsymbol{x}_{\boldsymbol{q}}^{k+1}) = \frac{1}{N}\sum_{i=1}^{N}f_{i}^{k+1}(\boldsymbol{x}_{\boldsymbol{q}}^{k+1}) \\
= (\boldsymbol{W}_0^{k+1}+\frac{1}{N^2} \sum_{i=1}^{N}\sum_{j=1}^{N} \Delta \boldsymbol{W}_{ji}^{k})\boldsymbol{q},
\end{aligned}
\label{eq:epoch}
\end{equation}

In this two-epoch setting, it's evident that \(N^{2}\) orders of 2-shot examples are contemplated in the summation of meta-gradients.
This implicit permutation not only maintains but also, to some extent, amplifies the interaction between examples, thereby enhancing the approach, all the while effectively preserving order-agnostic properties.

\subsection{Discussion on the Efficiency}
\label{sec:method-efficiency}
We focus on the computational overhead related to the attention mechanism in Transformer models. 
This part is a critical efficiency bottleneck for Transformers, showcasing a time complexity of \(O(n^2)\) when processing texts of length \(n\).

Consider inputs \((I_1, l_1, I_2, l_2, \dots, I_N, l_N, I_q)\), and suppose the average length of both an input and a label is \(T\).
In a standard \(N\)-shot ICL task, the total length of \(N\) examples combined with the query is \((2N+1)T\), resulting in an approximate time cost of \(O((2N+1)^{2}T^{2})\). 
In contrast, \name\ deconstructs the \(N\)-shot scenario into \(N\) separate 1-shot instances, which are subsequently integrated with a 0-shot learning. 
Specifically, we process \((I_i, l_i, I_q)\), each with a length of \(3T\), \(N\) times, and \((I_q)\) with a length of \(T\) once, culminating in a total time cost of approximately $10T^{2}$.

When \(N > 2\), it is obvious that \name\ outperforms the standard ICL in terms of efficiency. 
It's important to note that, in practical applications, we execute the $N$ 1-shot learning tasks simultaneously in a batch, which effectively reduces the actual latency. 
Furthermore, in many situations, particularly in classification tasks, the length of the label is considerably shorter than that of the input. 
Therefore, the benefits of our approach are even more pronounced.

To analyze practical resource utilization, we tested the AGnews dataset utilizing $N$=4 on an A100-80G. 
\name\ completes the inference of the entire dataset in 512.84 seconds, while the standard ICL method requires 697.09 seconds.
Considering space efficiency, \name\ provides $N$ times greater space efficiency than standard ICL under ideal conditions. 
In practice, executing the standard 4-shot AGnews task on Llama 2-13B consumes 32.1 GB of memory, while our method requires only 29.7 GB, wherein model parameters occupy 26 GB.

\section{Comparison between \name\ and standard $N$-shot ICL}
\label{sec:exp}

\subsection{Experimental Setup}
We conduct this preliminary study to assess \name\ in comparison to the standard $N$-shot ICL.
We choose widely-used open-source LLMs, Llama-2-7B and -13B~\cite{touvron2023llama}.

\begin{table*}[t]
\centering
\setlength{\tabcolsep}{5pt}
\resizebox{0.96\linewidth}{!}{\begin{tabular}{lcccccccccccc}
\toprule
\multirow{2}{*}{Task} &   \multicolumn{5}{c}{\textbf{7B}} &       \multicolumn{5}{c}{\textbf{13B}}  \\ 
\cmidrule(lr){2-6}
\cmidrule(lr){7-11}
 & $N$-shot & \name & Best &  PCW & F-Ordered & $N$-shot & \name & Best & PCW & F-Ordered \\
\midrule
\multirow{1}{*}{SST-2} & $54.9_{\hspace{0.05cm}\text{[6.4]}}$ & \hspace{0.2cm}$\textbf{58.7}_{\hspace{0.05cm}\text{[9.2]}}$ &\hspace{0.1cm} $58.8_{\hspace{0.05cm}[9.4]}$ & $54.2_{\hspace{0.05cm}[6.9]}$ & $56.4_{\hspace{0.05cm}[11.8]}$ & $63.8_{\hspace{0.05cm}[15.1]}$  & \hspace{0.3cm}$ \textbf{69.7}_{\hspace{0.05cm}[11.4]}$ & \hspace{0.3cm}$ 70.2_{\hspace{0.05cm}[11.5]}$ & $51.9_{\hspace{0.05cm}[3.7]}$& $67.7_{\hspace{0.05cm}[10.9]}$\\
\vspace{0.09cm} 
\multirow{1}{*}{RTE}  & $53.0_{\hspace{0.05cm}\text{[6.8]}}$ & \hspace{0.2cm}$\textbf{65.2}_{\hspace{0.05cm}\text{[1.9]}}$ &\hspace{0.1cm} $67.6_{\hspace{0.05cm}[0.5]}$  & $56.6_{\hspace{0.05cm}[5.2]}$ & $53.9_{\hspace{0.05cm}[5.1]}$& $78.1_{\hspace{0.05cm}[4.2]}$ & \hspace{0.3cm}$ \textbf{79.2}_{\hspace{0.05cm}[3.5]}$ & \hspace{0.3cm}$ 80.9_{\hspace{0.05cm}[2.8]}$ & $56.5_{\hspace{0.05cm}[7.3]}$& $80.5_{\hspace{0.05cm}[1.3]}$\\
\vspace{0.09cm} 
\multirow{1}{*}{QNLI}  & $50.7_{\hspace{0.05cm}\text{[1.8]}}$ & \hspace{0.2cm}$\textbf{52.4}_{\hspace{0.05cm}\text{[1.3]}}$ &\hspace{0.1cm} $55.2_{\hspace{0.05cm}[0.7]}$ & $50.1_{\hspace{0.05cm}[0.6]}$ & $51.3_{\hspace{0.05cm}[0.8]}$ & $51.1_{\hspace{0.05cm}[2.5]}$ & \hspace{0.3cm}$ \textbf{56.7}_{\hspace{0.05cm}[0.9]}$ & \hspace{0.3cm}$ 56.9_{\hspace{0.05cm}[1.2]}$ & $51.2_{\hspace{0.05cm}[1.1]}$& $56.2_{\hspace{0.05cm}[3.9]}$\\
\vspace{0.09cm}
\multirow{1}{*}{AGNews}  & $28.7_{\hspace{0.05cm}\text{[3.8]}}$ & \hspace{0.2cm}$29.3_{\hspace{0.05cm}\text{[3.1]}}$ &\hspace{0.1cm} $29.4_{\hspace{0.05cm}[3.0]}$ & $25.8_{\hspace{0.05cm}[2.0]}$& $\textbf{30.2}_{\hspace{0.05cm}\text{[4.9]}}$ & $30.5_{\hspace{0.05cm}[6.3]}$ & \hspace{0.3cm}$ 31.4_{\hspace{0.05cm}[3.9]}$ & \hspace{0.3cm}$ 31.6_{\hspace{0.05cm}[3.8]}$ & $25.5_{\hspace{0.05cm}[0.8]}$& $\textbf{32.1}_{\hspace{0.05cm}[6.5]}$\\
\vspace{0.09cm}
\multirow{1}{*}{MRPC}  & $40.5_{\hspace{0.05cm}\text{[10.5]}}$ & \hspace{0.2cm}$\textbf{66.9}_{\hspace{0.05cm}\text{[0.5]}}$ &\hspace{0.1cm} $68.0_{\hspace{0.05cm}[0.2]}$ & $56.8_{\hspace{0.05cm}[14.3]}$ & $38.6_{\hspace{0.05cm}\text{[7.6]}}$& $52.4_{\hspace{0.05cm}[14.8]}$ & \hspace{0.3cm}$ 56.5_{\hspace{0.05cm}[0.6]}$ & \hspace{0.3cm}$ 67.9_{\hspace{0.05cm}[1.7]}$ & $\textbf{63.8}_{\hspace{0.05cm}[8.0]}$& $63.3_{\hspace{0.05cm}[6.5]}$\\
\bottomrule
\end{tabular}}
\caption{Experimental results of classification tasks. 
We report the average score and standard deviation [$\sigma$] across 10 runs. 
The ``Best'' column reports the upper limit of \name\ in which we search for the optimal $k$ for each test sample.
We also compare \name\ with PCW~\cite{ratner-etal-2023-parallel} and Fantastically Ordered~\cite{lu-etal-2022-fantastically}.
\name\ not only enhances LLMs' ICL performance in diverse tasks but also decreases performance variability across different demonstration examples, often approaching its maximum potential.
All bold results passed the t-test with a p-value < 0.05. }
\label{table:main_results}
\end{table*}

\begin{table}[t]
    \centering
    \begin{tabular}{lcc}
\toprule
& $N$-shot & \name \\\midrule
7B &    $66.38$ & \hspace{0.1cm}$65.91$\\
13B &    $67.88$ & \hspace{0.1cm}$67.63$\\\bottomrule
    \end{tabular}
    \caption{BLEU scores for WMT2014 En-Fr.}
    \label{tab:wmt}
\end{table}

We evaluate \name\ on classification tasks, containing one sentiment detection dataset \textbf{SST-2} \cite{socher-etal-2013-recursive}, two natural language inference datasets \textbf{RTE} \cite{10.1007/11736790_9, NEURIPS2019_4496bf24} and \textbf{QNLI} \cite{wang-etal-2018-glue}, a topic classification dataset \textbf{AGNews} \cite{NIPS2015_250cf8b5} and one paraphrase dataset \textbf{MRPC} \cite{dolan-brockett-2005-automatically}. 
Moreover, we also utilize \name\ for free-format generation tasks, such as machine translation. 
We choose the \textbf{WMT2014 En-Fr}~\citep{bojar-etal-2014-findings} benchmark.

Additionally, we compare \name\ with Parallel Context Windows (PCW,~\citealp{ratner-etal-2023-parallel}) that also enhances the inference of ICL. 
In terms of parallel processed examples, PCW is the most related work to our study. 
PCW resets the position embedding for each example, implicitly handling these examples in parallel. 
However, it does not improve efficiency and introduces gaps in inference behavior, potentially compromising performance.
We also compare \name\ with Fantastically Ordered (F-Ordered,~\citealp{lu-etal-2022-fantastically}), a prior study on selecting the optimal order for ICL. 
F-Ordered necessitates the enumeration of all permutations to identify an optimal prompt sequence.

For classification tasks, we evaluate the accuracy on the test set when labels are accessible, and we sample 300 data from the validation set to derive the optimal value of $k$.
In cases where test labels are not provided, we evaluate performance on the validation set and choose 300 data from the training set for obtaining $k$.
For the machine translation task, we determine the optimal value of $k$ using the validation set and report the BLEU score~\cite{papineni-etal-2002-bleu}.

In this section, we fix the value of $N$ as 4.
Unless specifically stated otherwise, we present the average score obtained from 10 runs, each with different sampled demonstration examples.
This ensures reliable conclusions, as different ICL demonstrations lead to varying performance~\citep{pmlr-v139-zhao21c, perez2021true}.

\subsection{Experimental Results}
We present the experimental results on classification tasks in Table~\ref{table:main_results}. 
Overall, \name\ demonstrates a substantial improvement over both the standard $N$-shot ICL and PCW in various tasks and across different model sizes. 
Across four out of the five datasets analyzed, \name\ surpasses the performance of F-Ordered, with marginal differences observed on the AGnews dataset.

To determine \name's maximum potential, we identified the optimal $k$ for each test sample and have included these findings in the ``Best'' column. 
These results show that the performance discrepancy between our chosen $k$ and the theoretical upper limit of performance is negligible. 
Furthermore, because \name\ employs a larger batch size during meta-optimization, which diminishes noise and randomness in the meta-gradients, we observe \name\ exhibits markedly more stable performance than standard ICL across different demonstrations, as evidenced by the significantly reduced variation\footnote{An exception occurs with Llama-2-7B in the SST-2 task, where \name\ increases variance.
This happens because, in this binary classification task, the ``$N$-shot'' is close to stable random guessing.} in repeated experiments.
These results serve as initial validation for the effectiveness of \name, and also support our explanation in Section~\ref{sec:method-explain}.

Table~\ref{tab:wmt} illustrates the performance comparison for the machine translation task. 
\name\ outperforms the standard ICL in efficiency.
It achieves this while preserving the BLEU score, experiencing only a minimal decrease of 0.7\% for the 7B model and 0.3\% for the 13B model, respectively.

To showcase the robustness and versatility of \name, we have implemented it across different LLMs, including OPT-6.7B~\cite{zhang2022opt} and Falcon-7B~\cite{almazrouei2023falcon}, using SST-2 as a benchmark task. Table~\ref{tab:llms} illustrates the effectiveness of \name\ across various models, demonstrating its universality.

\begin{table}[t]
    \centering
    \begin{tabular}{lcc}
\toprule
& $N$-shot & \name \\\midrule
	OPT-6.7B	 &    $38.4$ & \hspace{0.2cm}$42.5$\\
Falcon-7B &    $51.2$ & \hspace{0.2cm}$52.0$\\\bottomrule
    \end{tabular}
    \caption{Experimental results of SST-2 on a broader array of LLMs.}
    \label{tab:llms}
\end{table}

\section{Method Analysis}
\label{sec:ablation}
We delve into various aspects of \name, including the impact of shot number ($N$), aggregation layer index ($k$), the impact of the order of ICL examples, and the optimization ``epochs.''
This exploration aims to provide a deeper insight into both the efficacy of \name\ and the ICL itself.

\subsection{The Effect of $N$}
\begin{figure}[t]
	\centering
	\subfigure[SST-2]{
		\begin{minipage}{0.8\linewidth} 
                        \centering
                        \includegraphics[width=\linewidth]{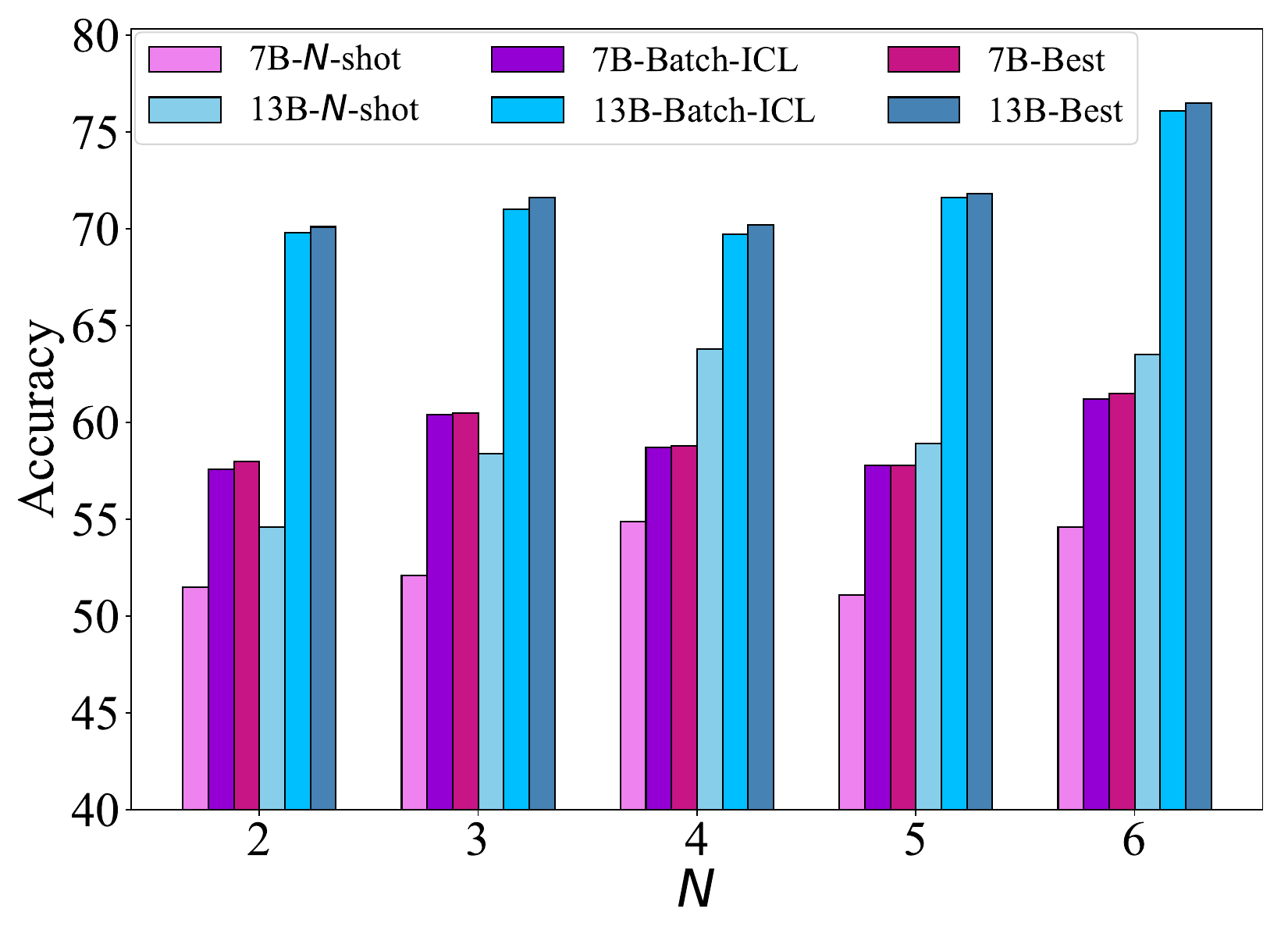} \\
		\end{minipage}
	}
	\subfigure[RTE]{
		\begin{minipage}{0.8\linewidth}
            \centering
			\includegraphics[width=\linewidth]{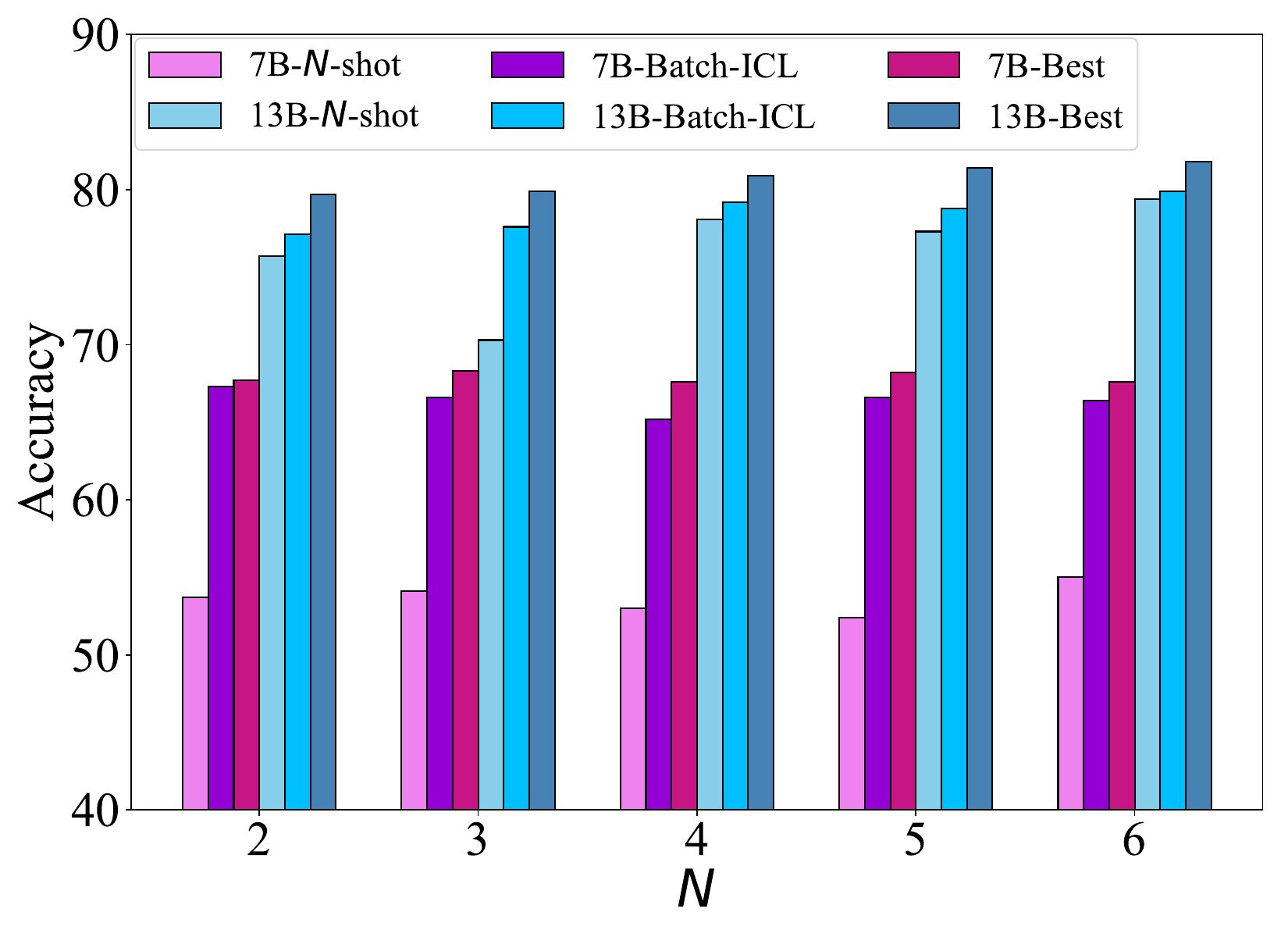} \\
			
		\end{minipage}
	}
    \caption{Performance dynamics across various $N$ on SST-2 and RTE.}
	\label{fig:ablationN}
\end{figure}
Different tasks typically demand varying values of $N$. 
The effectiveness of \name\ across different $N$ represents its robustness in various practical applications. 
As illustrated in Figure~\ref{fig:ablationN}, our \name\ consistently surpasses the standard $N$-shot accuracy for different values of $N$. 
Also, the gap between the performance of \name\ and its upper limit referred to as ``Best'' is minor and diminishes as the value of $N$ increases.

\begin{table}[t]
\centering
\resizebox{0.9\linewidth}{!}{
\begin{tabular}{ccccc}
\hline
    &  $N=4$ &  $N=10$&  $N=20$&  $N=70$ \\ \hline
7B   & 29.3  & 36.24& 39.43& 33.3 \\
13B & 31.4 & 32.67	&33.70&39.1 \\ \hline
\end{tabular}}
\caption{Results from \name\ on the AGNews task with varying values of $N$.
Even when $N=70$, demonstration examples surpass the model's maximum context length, yet it still achieves additional improvements.}
\label{table:length}
\end{table}

\begin{figure}[t]
    \centering
    \includegraphics[width=0.92\linewidth]{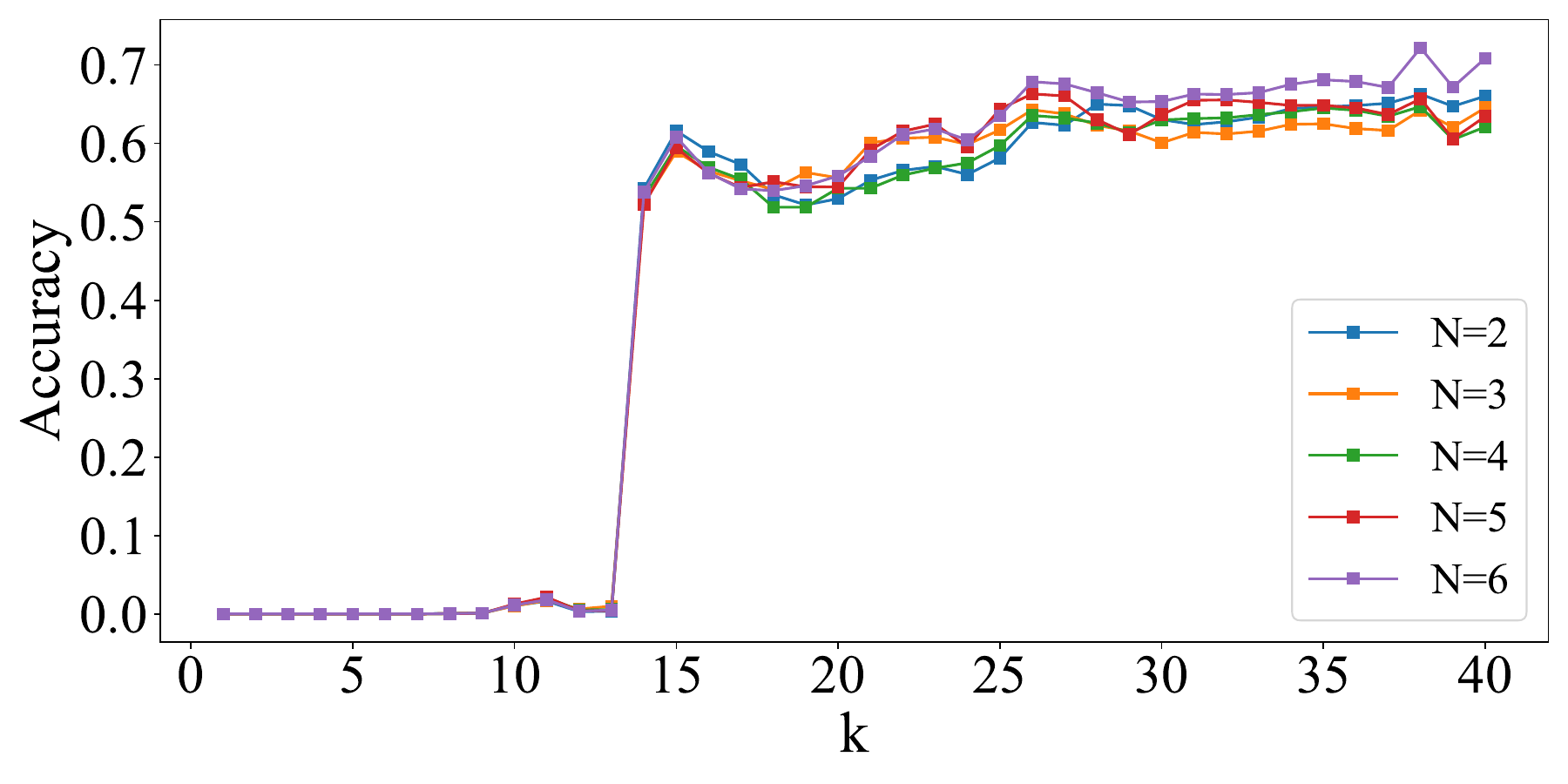}
    \caption{Performance dynamics across various aggregation layer $k$.}
    \label{fig:ablation-k}
\end{figure}

\begin{figure}[t]
    \centering
    \includegraphics[width=0.8\linewidth]{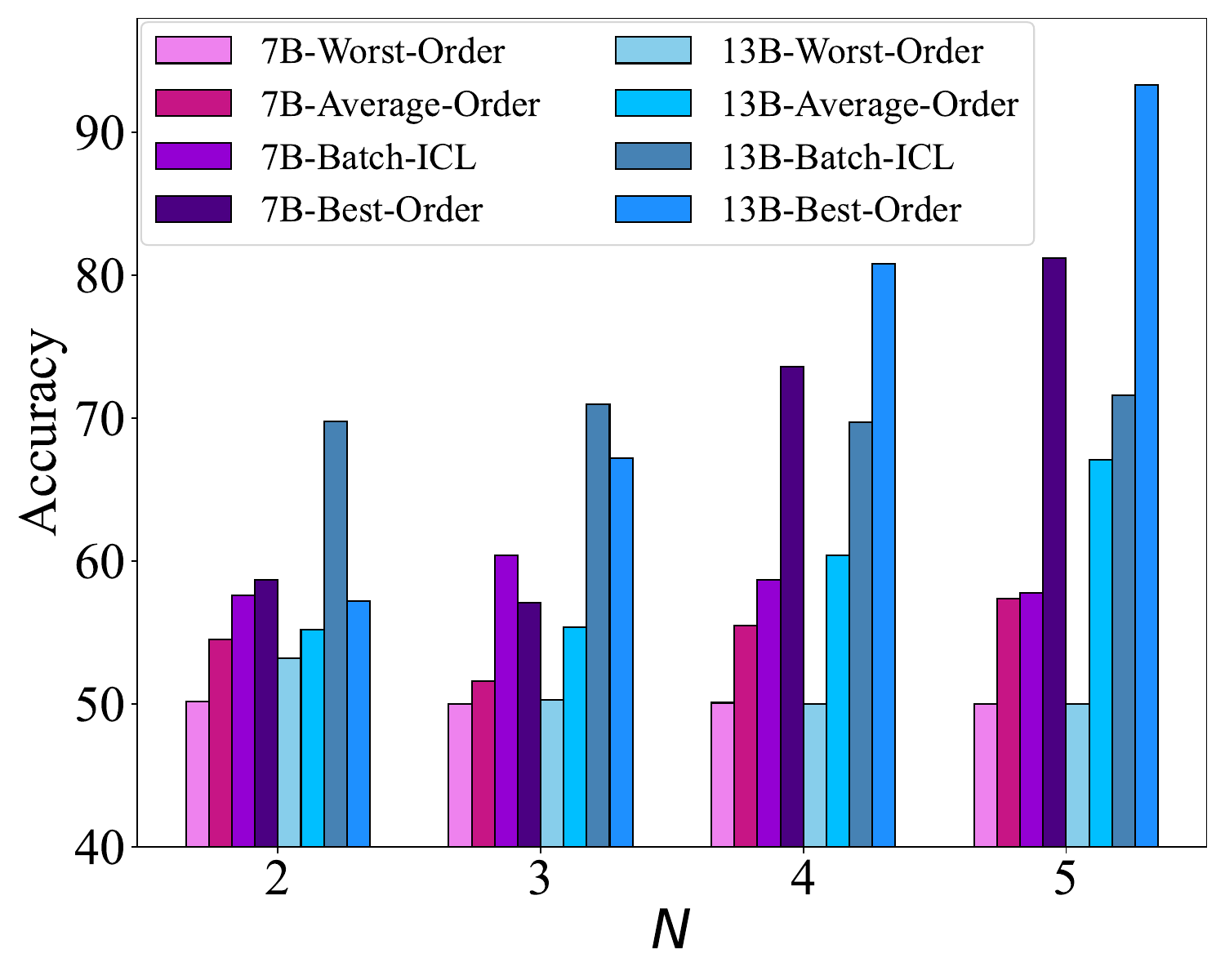}
    \caption{
    Comparing \name\ and standard ICL with various example orders, including the ``Best'', ``Worst'' and ``Average'' of all permutations.
    }
    \label{fig:order}
\end{figure}

\begin{table}[t]
\centering
\resizebox{0.8\linewidth}{!}{
\begin{tabular}{l|cccc}
\hline
 \diagbox{Model}{$N$}& 2 & 3 & 4 & 5 \\ \hline
 Llama-2-7B  & $60.0$ & $90.0$  & $65.4$ & $63.7$ \\
 Llama-2-13B & $80.0$ & $88.3$ & $71.3$ & $60.5$ \\ \hline
\end{tabular}}
\caption{The percentage of permutations that \name\ outperforms.}
\label{table:ord}
\end{table}

Due to the parallel processing of ICL examples, as long as each individual example doesn't exceed the maximum context length, there's no limit to the number of examples \name\ can handle.
In Table~\ref{table:length}, for the AGNews task, as $N$ increases from $4$ to $70$, we found the performance first hits a peak and then drops on Llama-2-7B; By contrast, \name\ keeps bringing benefits with a larger $N$ on Llama-2-13B. 
We assume the drop in performance at $N=70$ on Llama-2-7B may stem from the fact that 7B model has already exploited its ability, while 13B is stronger at learning from more shots.
Notably, even when we include $N=70$ examples, averaging a total of 5,447 tokens, which significantly exceeds Llama-2's context limit of 4,096, \name\ continues to demonstrate additional enhancements.

These analyses demonstrate that \name\ is robust to $N$ and can more thoroughly exploits LLMs' in-context learning capacity.

\begin{table*}[t]
\centering
\resizebox{0.8\linewidth}{!}{
\begin{tabular}{lcccccccccc}
\toprule
\multirow{2}{*}{N} & \multicolumn{5}{c}{\textbf{7B}} & \multicolumn{5}{c}{\textbf{13B}} \\ 
\cmidrule(lr){2-6}
\cmidrule(lr){7-11}
 & \textbf{Epoch} & \textbf{1} & \textbf{2} & \textbf{3} & \textbf{4} & \textbf{Epoch} & \textbf{1} & \textbf{2} & \textbf{3} & \textbf{4} \\
\midrule
\multirow{2}{*}{\centering 2} & \textbf{SST-2} & $57.97$ & $58.78$ & $58.91$ & $\textbf{59.48}$ & \textbf{SST-2} & $70.09$ & $\textbf{70.30}$ & $69.79$ & $69.83$ \\
 & \textbf{RTE} & $67.7$ & $\textbf{67.8}$ & $67.7$ & $67.9$ & \textbf{RTE} & $79.7$ & $\textbf{80.0}$ & $79.5$ & $79.7$ \\
\multirow{2}{*}{\centering 3} & \textbf{SST-2} & $60.49$ & $64.27$ & $65.11$ & $\textbf{67.77}$ & \textbf{SST-2} & $\textbf{71.58}$ & $71.35$ & $71.39$ & $71.40$ \\
 & \textbf{RTE} & $68.2$ & $\textbf{68.3}$ & $68.3$ & $68.3$ & \textbf{RTE} & $79.9$ & $\textbf{80.1}$ & $80.0$ & $80.0$ \\
\multirow{2}{*}{\centering 4} & \textbf{SST-2} & $58.84$ & $60.52$ & $60.60$ & $\textbf{62.10}$ & \textbf{SST-2} & $\textbf{70.19}$ & $69.90$ & $69.93$ & $69.83$ \\
 & \textbf{RTE} & $67.6$ & $\textbf{67.9}$ & $67.8$ & $67.7$ & \textbf{RTE} & $80.9$ & $80.9$ & $81.1$ & $\textbf{81.3}$ \\
\multirow{2}{*}{\centering 5} & \textbf{SST-2} & $57.80$ & $59.96$ & $60.32$ & $\textbf{61.39}$ & \textbf{SST-2} & $71.80$ & $\textbf{71.82}$ & $71.37$ & $71.26$ \\
 & \textbf{RTE} & $68.2$ & $68.3$ & $\textbf{68.4}$ & $68.1$ & \textbf{RTE} & $81.4$ & $\textbf{81.6}$ & $81.4$ & $81.4$ \\
\multirow{2}{*}{\centering 6} & \textbf{SST-2} & $61.49$ & $63.08$ & $63.32$ & $\textbf{64.57}$ & \textbf{SST-2} & $76.52$ & $\textbf{76.55}$ & $76.26$ & $75.84$ \\
 & \textbf{RTE} & $67.6$ & $67.7$ & $67.8$ & $\textbf{67.9}$ & \textbf{RTE} & $81.8$ & $\textbf{82.0}$ & $82.0$ & $82.0$ \\
\bottomrule
\end{tabular}
}
\caption{Multiple epochs prove to enhance accuracy on SST-2 and RTE, irrespective of model size and shots. Note that, in this study, we do not determine $k$ based on the validation set but rather employ the optimal value for each individual sample.}
\label{table:epochs}
\end{table*}

\subsection{The Effect of $k$}

We examined the effects of $k$ by varying its value across the full range of layers, monitoring the performance dynamics as aggregated meta-gradients are implemented at various levels. 
For each value of $k$, we carried out experiments using the Llama-2-13B model on the SST-2 dataset~\citep{socher-etal-2013-recursive}, with $N$ values ranging from 2 to 6.

Figure~\ref{fig:ablation-k} demonstrates that with a small $k$, \name\ does not work. 
It suggests that the initial, or shallow, layers of LLMs play a crucial role in establishing semantic foundations for the subsequent, deeper layers. 
This observation aligns with previous studies~\citep{wang-etal-2023-label, todd2023function, hendel2023incontext}.

As the forward computation goes on, we observe an abrupt increase in performance. 
This improvement plateaus rapidly and does not diminish thereafter. 
This pattern differs from the findings reported in~\citep{hendel2023incontext, todd2023function}, where researchers found that a deeper representation of an ICL task diminishes the task-related information it contains. 
Specifically, when the representation is extracted and used in a zero-shot forward computation, the accuracy tends to converge to zero as the layer depth increases.
The discrepancy between ours and theirs findings might be due to subtle differences in input configuration. 
In our approach, each 1-shot forward computation uses a true query. 
In contrast, previous studies used a pseudo query, which can lead to significant deviations in deeper layers because most information gathers at the last position~\citep{wang-etal-2023-label}.

\subsection{The Effect of ICL Example Order}
We meticulously examine all $N!$ possible permutations of ICL examples in a $N$-shot SST-2~\citep{socher-etal-2013-recursive} task. 
Due to the prohibitively high cost associated with enumerating the permutations, we restrict the value of $N$ within the range of 2 to 5.
In Figure~\ref{fig:order}, the ``Average'' result is the average accuracy of all permutations. 
``Best Order'' denotes the highest accuracy, contrasting with the ``Worst Order'' which represents the lowest. 

It is clear that, regardless of the model's size, the performance under the worst order is close to a random guess.
In stark contrast, the best order significantly surpasses the average performance. 
This emphasizes the critical role of sequence organization in in-context learning~\citep{lu-etal-2022-fantastically}.

In situations where only a few examples are available (e.g., $N$=2 or 3), \name\ not only exceeds the average performance but also surpasses the best order. 
This underscores the strength of our proposed method in data-limited scenarios. 
As the number of examples increases, the number of permutations rises, enhancing the likelihood of achieving high accuracy; consequently, the best-order performance improves markedly. 
Nevertheless, \name\ consistently demonstrates superior performance compared to the average.

Additionally, we performed a statistical analysis on the percentage of permutations that \name\ outperforms. 
In Table~\ref{table:ord}, it is obvious that \name\ exceeds the majority of permutations for every $N$. 
This is particularly noticeable for a smaller $N$ and a larger model.
Overall, this study confirms our motivation, demonstrating that \name\ effectively alleviates concerns related to ICL example orders, leading to a satisfactory solution.

\subsection{The Effect of Epochs}
In Table \ref{table:epochs}, we present the results of adding more epochs to \name\ on SST-2~\citep{socher-etal-2013-recursive} and RTE~\cite{10.1007/11736790_9, NEURIPS2019_4496bf24}. 
This table illustrates the model's performance across various numbers of $N$ and a range of epochs. 
Our findings indicate that, across the majority of $N$ and model sizes, there is an improvement in the model's performance over several epochs.
Specifically, in the RTE task, we observe a performance plateau typically achieved within 2 epochs. 
Conversely, in the SST-2 task, we extend our investigation across more epochs, typically reaching a plateau after approximately 10 to 20 epochs, as illustrated in Table \ref{table:moreepo}.

\begin{table}[t]
\centering
\begin{tabular}{cccc}
\hline
    &  Epoch=4 & Epoch=10 & Epoch=20 \\ \hline
7B   & 62.10  & 67.84 & 67.40 \\
13B & 69.83 & 76.30 & 72.38\\ \hline
\end{tabular}
\caption{Performance of \name\ with more epochs.}
\label{table:moreepo}
\end{table}

Overall, we observed a more pronounced effect in the 7B model compared to the 13B model. 
The 7B model consistently shows enhanced performance with an increasing number of epochs, peaking at 4 epochs. 
On the other hand, the 13B model reaches the plateau earlier. 
This could be due to the larger capacity of the 13B model, which enables it to more effectively capture linguistic subtleties and contextual details with fewer implicit interactions among ICL examples (see Eq.\ref{eq:epoch}).

\section{Related Works}
\label{sec:relate}
In addition to PCW we compared in \S\ref{sec:exp}, our study is also related to the findings of \citet{hendel2023incontext} and \citet{todd2023function}. 
These works discovered that in few-shot learning tasks, the representation in deep layers carries the task information.
\citet{hendel2023incontext} focused on identifying and aggregating the outputs of induction heads~\citep{olsson2022context}. 
They emphasized the content analysis of aggregated hidden states. 
In contrast, we focus on a simpler yet effective method for enhancing the ICL and do not differentiate specific heads.
When compared to~\citet{hendel2023incontext}, our method demonstrates greater efficacy and surpasses the original few-shot learning performance by using zero-shot learning, an achievement not realized by~\citet{hendel2023incontext}.
Theoretically, our work offers an explanation for the ``training set compression'' proposed by~\citet{hendel2023incontext} and elucidates why averaged attention activation conveys ICL task information~\citep{todd2023function}.

\section{Conclusion}
\label{sec:conclusion}
From a meta-optimization perspective, we explain ICL example orders' impact on performance. 
Building on our insights, we introduce \name, an efficient and effective algorithm for ICL inference. 
\name\ processes ICL examples in batches, aggregating their meta-gradients.
Aggregated meta-gradients are then applied to a zero-shot forward computation for final predictions. 
Due to the batch processing, \name\ is agnostic to the order of ICL examples, surpassing the average performance of all order permutations across various tasks, and supports much more examples.
We expand \name\ by developing multi-epoch variants that implicitly enumerate permutations of ICL examples, which fosters better interaction between inputs and further improves our method.

\section*{Acknowledgments}
This work was supported by the National Natural Science Foundation of China (NSFC Grant No. 62122089), Beijing Outstanding Young Scientist Program NO. BJJWZYJH012019100020098, and Intelligent Social Governance Platform, Major Innovation \& Planning Interdisciplinary Platform for the "Double-First Class" Initiative, Renmin University of China, the Fundamental Research Funds for the Central Universities, and the Research Funds of Renmin University of China.
This work was supported by Ant Group Research Fund.
Ang Lv is supported by the Outstanding Innovative Talents Cultivation Funded Programs 2023 of Renmin University of
China.

\newpage
\section*{Limitations}
The theoretical foundation of our work is grounded in various studies ~\citep{1964,Irie2022TheDF,dai-etal-2023-gpt} which takes the attention in Transformers as a linear attention.
Some research~\cite{ren2023incontext} suggests that even without this simplification, the conclusions and insights of these studies remain valid.
Nevertheless, for the sake of clarity in presentation and ease of comprehension, we adhere to the linear simplification. 
Additionally, in the section where we explore multiple ``epochs,'' we simplify the feed-forward layer as a linear transformation. 
This simplification is widely adopted in many works on interpretability~\cite{olsson2022context,wang2023interpretability,yu2023characterizing,wang2023label} due to the considerable challenges associated with analyzing nonlinear MLP.

The potential risks of our study are similar to those of other works involving LLMs, as they can sometimes generate toxic responses.
\bibliography{custom}
\bibliographystyle{acl_natbib}

\end{document}